\documentclass{article}

\usepackage{arxiv}

\usepackage{times}
\usepackage{latexsym}
\usepackage[T1]{fontenc}
\usepackage[utf8]{inputenc}
\usepackage{microtype}
\usepackage{inconsolata}
\usepackage{tabularx}
\usepackage{graphicx}
\usepackage{booktabs}
\usepackage{multirow}
\usepackage{amsmath}
\usepackage{amssymb}
\usepackage{xcolor}
\usepackage{colortbl}
\usepackage{xspace}
\usepackage{tikz}
\usetikzlibrary{positioning, calc, arrows.meta, fit, backgrounds}
\usepackage{tcolorbox}
\tcbuselibrary{skins}
\usepackage{placeins}
\usepackage{algorithm}
\usepackage{algpseudocode}
\usepackage{epigraph}
\renewcommand{\sourceflush}{flushright}

\usepackage{url}
\usepackage{natbib}
\usepackage[colorlinks=true, linkcolor=rmBlue, citecolor=rmBlue, urlcolor=rmBlue]{hyperref}
\newcommand{\kbfqa}{\textsc{KBF-QA}\xspace}
\newcommand{\ours}{\textsc{ReverieMem}\xspace}
\newcommand{\bookworld}{\textsc{BookWorld}\xspace}
\newcommand{\kbf}{\textsc{KBF}\xspace}
\newcommand{\krf}{\textsc{KRf}\xspace}
\newcommand{\kr}{\textsc{KR}\xspace}

\definecolor{rmBlue}{RGB}{45,85,160}       %
\definecolor{rmGray}{RGB}{90,90,90}        %
\definecolor{rmGreen}{RGB}{50,115,75}      %
\definecolor{rmViolet}{RGB}{105,65,145}    %
\definecolor{rmOrange}{RGB}{200,120,30}    %
\definecolor{rmRed}{RGB}{170,55,75}        %

\newcommand{\rmphase}[2]{%
  \tcbox[
    enhanced, on line, nobeforeafter,
    colback=#1!8, colframe=#1!85!black,
    arc=2pt, boxrule=0.4pt,
    boxsep=0pt, top=0.5pt, bottom=0.5pt, left=2pt, right=2pt,
    fontupper=\footnotesize\bfseries\color{#1!85!black}
  ]{\strut{\scriptsize #2}}%
}

\newcommand{\rmlayer}[2]{\textbf{\textcolor{#1!85!black}{#2}}}

\title{Staying In Character: Role-Playing Agents with Perspective-Bounded Memory}
\date{}
\renewcommand{\headeright}{}
\renewcommand{\undertitle}{}

\author{
  Xushuo Tang\textsuperscript{1}\thanks{Equal contribution.} \quad
  Junhe Zhang\textsuperscript{1}\footnotemark[\value{footnote}] \quad
  Zihan Yang\textsuperscript{3} \quad
  Yifu Tang\textsuperscript{4} \\
  \bfseries Sichao Li\textsuperscript{2} \quad
  Longbin Lai\textsuperscript{5} \quad
  Zhengyi Yang\textsuperscript{2} \\[6pt]
  \normalfont\small\textsuperscript{1}UNSW Sydney \quad
  \textsuperscript{2}University of Sydney \quad
  \textsuperscript{3}Chang'an University \\
  \textsuperscript{4}RAIDS Lab \quad
  \textsuperscript{5}Tongyi Lab, Alibaba Group \\[4pt]
  \texttt{xushuo.tang@unsw.edu.au, junhe.zhang1@student.unsw.edu.au, zihany@chd.edu.cn,} \\
  \texttt{yves.tang@raids-lab.com, sichao.li@sydney.edu.au, zhengyi.yang@sydney.edu.au,} \\
  \texttt{Longbin.lailb@alibaba-inc.com}
}

\begin{document}
\maketitle

\begin{abstract}
Recent LLM role-playing systems build character agents from novels by extracting characters, scenes, and relations.
Yet long-narrative role playing suffers from two failures: \emph{Factual Overreach}, where shared retrieval or parametric memory lets a character use facts outside its perspective, and \emph{Stylistic Monotony}, where profile descriptions flatten a character into a fixed voice.
To address these failures, we propose \ours, a three-layer memory architecture for book-based character agents.
The episodic layer stores first-person scene memories; the semantic layer stores visibility-tagged facts; and the personality layer stores situation-dependent speech and behaviour patterns.
We evaluate \ours and six reference methods on \kbfqa, a new benchmark of 4{,}386 questions across eight novels that assesses character knowledge boundaries.
\ours improves Knowledge Boundary Fidelity by 34.6 percentage points over the strongest reference method.
On \bookworld's five-dimension pairwise narrative protocol, \ours achieves a mean win rate of 78.86\% against \bookworld across five generator LLMs and two generation modes, suggesting that perspective-bounded memory improves both boundary fidelity and character-grounded narrative generation.
\end{abstract}

\section{Introduction}
\label{sec:intro}

\begingroup
\renewcommand{\sourceflush}{flushright}
\epigraph{%
   ``The divine gift does not come from a higher power, but from our own minds.''%
}{--- Dr.\ Robert Ford, \emph{Westworld}}
\endgroup

\begin{figure}[!t]
   \centering
   \includegraphics[width=0.9\textwidth]{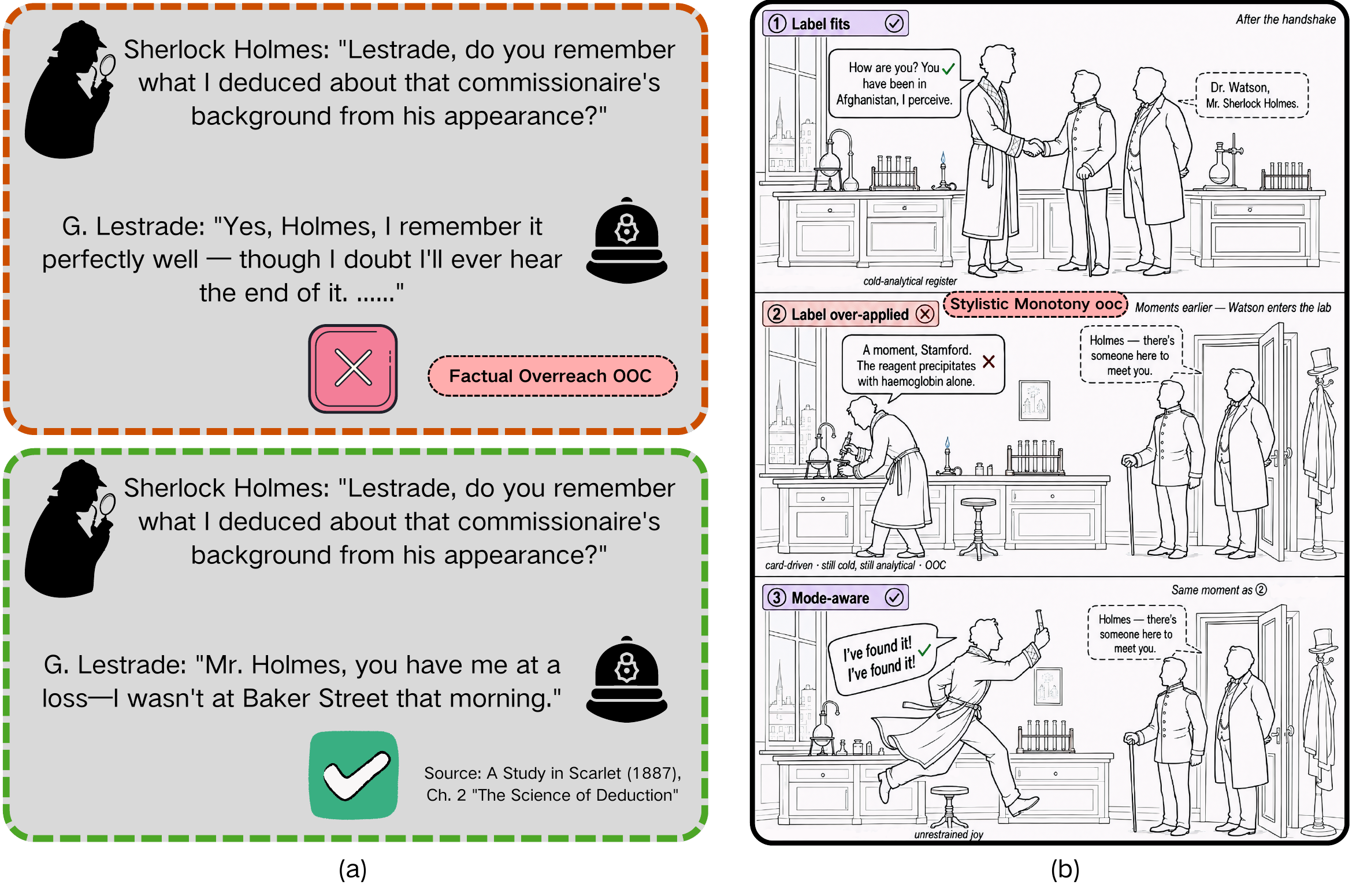}
   \caption{Two OOC failures in long-narrative role playing: \textbf{Factual Overreach} (a), where a character claims facts outside its perspective, and \textbf{Stylistic Monotony} (b), where a character's expression is flattened into one fixed profile.}
   \label{fig:teaser}
\end{figure}

LLM role-playing agents have become a practical interface for character chat, narrative sandboxes, and interactive story generation~\citep{shao-etal-2023-character,wang-etal-2024-rolellm,charactereval,hollmwood}.
Recent work builds such agents from novels by extracting characters, scenes, and relations as narrative context~\citep{zhao-etal-2024-narrativeplay,bookworld,wang-etal-2025-coser}; for example, \bookworld constructs interactive agent societies from books and guides characters with profiles and story context~\citep{bookworld}.
A credible character agent must reason within the character's knowledge boundary and dynamically adapt its voice and behaviour to the evolving narrative situation.
Role-playing agents face two OOC (out-of-character) failures, illustrated in Figure~\ref{fig:teaser}: \textbf{Factual Overreach} and \textbf{Stylistic Monotony}.

The first failure, \textbf{Factual Overreach}, occurs when an agent states canonically true facts that are unavailable to the character it is playing.
This can happen when long-narrative RAG systems  retrieve from a shared book-level memory, or when the LLM's parametric memory supplies the fact directly~\citep{hipporag,comorag2026}.
In Figure~\ref{fig:teaser} (a), Lestrade should not confirm Holmes's deduction about the commissionaire's background at Baker Street: the deduction is correct, but Lestrade was not present.

The second failure, \textbf{Stylistic Monotony}, occurs when profiles collapse a character into one expressive mode. In Figure~\ref{fig:teaser} (b),
Holmes, for example, is not only a cold analytical reasoner: in Doyle's first chapter, he greets Watson as an excited experimenter (\textit{``I've found it!''}), then moments later becomes the composed deducer who infers Watson's military past (\textit{``Afghanistan, I perceive''}).
A static profile may produce a recognisable Holmes while losing this situational shift.

These failures motivate a memory design that scopes factual access to the character's narrative position and grounds expression in situated behaviour across scenes.

Three cognitive accounts guide the design: 1) \emph{Complementary Learning Systems}~\citep{cls1995} models memory as cooperation between hippocampal encoding of individual experiences and neocortical consolidation of structured knowledge; 2) Conway's \emph{Self-Memory System}~\citep{conway2000self} casts autobiographical recall as a top-down reconstruction that locates event context first and reconstructs details within it; 3) \emph{Narrative identity}~\citep{mcadams2013} holds that identity is constituted by self-defining episodes rather than static traits.

We therefore present \ours, a cognitive-psychology-inspired three-layer memory architecture for role-playing agents.
The \emph{Episodic Layer} stores first-person scene summaries from each character's perspective; the \emph{Semantic Layer} stores structured facts with character-specific visibility; and the \emph{Personality Layer} stores situation-dependent patterns distilled from canonical speech, behaviour, and emotion transitions.
At inference time, \ours anchors on the character's own scenes, retrieves only visibility-allowed facts, and conditions the response on an appropriate behavioural pattern.

Novels record characters' experiences and behaviour across situations, providing a basis for evaluating these requirements. We evaluate \ours on eight novels with two complementary tasks.
For knowledge boundaries, we construct \kbfqa(Knowledge Boundary Fidelity-QA), a 4{,}386-question multiple-choice benchmark in which a character must answer facts it could know and refuse facts outside its perspective.
For open-ended narrative quality, we follow the five-dimension pairwise comparison protocol of \bookworld.
\ours improves KBF by 34.6 percentage points over the strongest prior method and achieves a mean win rate of 78.86\% against \bookworld in pairwise narrative comparisons.

This work makes three contributions:

\begingroup
\setlength\leftmargini{1.2em}
\begin{itemize}\setlength\itemsep{1pt}
\item We jointly address two OOC failure modes in role-playing LLM agents, \emph{Factual Overreach} and \emph{Stylistic Monotony}, by treating character knowledge access and situation-dependent expression as complementary design objectives.
\item We propose \ours, a three-layer memory architecture that separates scene experience, visibility-gated factual knowledge, and situation-dependent expression for role-playing agents.
\item We construct \kbfqa to evaluate character-specific recall and refusal, and show that \ours outperforms all six baselines on knowledge-boundary fidelity and \bookworld in pairwise narrative comparisons.
\end{itemize}
\endgroup

\section{Related Work}

\subsection{Role-Playing Agents}

Role-playing agents test whether LLMs can sustain consistent characters in dialogue and narrative tasks. One line of work improves role consistency through persona descriptions, dialogue histories, source-text grounding, synthetic personas, and role-playing training~\citep{shao-etal-2023-character,wang-etal-2024-rolellm,wang-etal-2025-opencharacter}, with literary settings further grounding agents in book scenes and character internals~\citep{wang-etal-2025-coser}. Benchmarks accompany this line and evaluate multi-turn dialogue, personality fidelity, and emotional fidelity~\citep{charactereval,wang-etal-2024-incharacter,feng-etal-2025-emocharacter}, while a recent survey separates fictional role playing from personalization~\citep{tseng-etal-2024-two-tales}. These works mainly test whether responses match the target persona, but provide less evidence on whether the character should know the expressed fact.

A second line builds story worlds populated by interacting agents through sandbox simulations, multi-agent frameworks, and novel-to-simulation systems~\citep{park-etal-2023-generative-agents,chen-etal-2023-agentverse,bookworld}, alongside multi-agent narrative-generation systems for long-story writing, screenwriting via role-played authoring, and autonomous plot progression~\citep{xia-etal-2025-storywriter,hollmwood,zhao-etal-2025-rolearena}. Characters in these systems are typically represented through structured profiles, relation states, or shared narrative context, and evaluation focuses on plot coherence and interaction quality rather than the speaking character's access to facts in the source narrative.

\subsection{RAG \& Memory for Narrative Reasoning}

Memory determines what evidence is available during generation. Retrieval-augmented generation grounds outputs in external documents~\citep{lewis-etal-2020-rag}, and adaptive retrieval, hierarchical summarisation, long-term memory, and narrative memory systems improve evidence selection and multi-hop or long-story reasoning~\citep{asai-etal-2024-self-rag,sarthi2024raptor,comorag2026}. For role-playing agents, however, the question is not only whether a retrieved fact is relevant but also whether the character can access it.

A complementary line addresses the boundary of what an agent should commit to. General LLM knowledge boundaries are catalogued in a recent survey~\citep{li-etal-2025-knowledge-boundary}; for role-playing agents specifically, boundary-aware training~\citep{tang-etal-2024-erabal} and representation-level refusal editing~\citep{liu-etal-2025-tell} aim to suppress out-of-role answers. TimeChara~\citep{ahn-etal-2024-timechara} evaluates point-in-time character hallucination and uses temporal and spatial experts to guide responses. Our memory architecture coordinates scene-anchored reasoning, character--fact visibility, and situation-dependent expression, coupling control over factual access with source-grounded behavioural guidance.

 \section{\ours}
\label{sec:method}

\begin{figure}[!t]
   \centering
   \includegraphics[width=\textwidth]{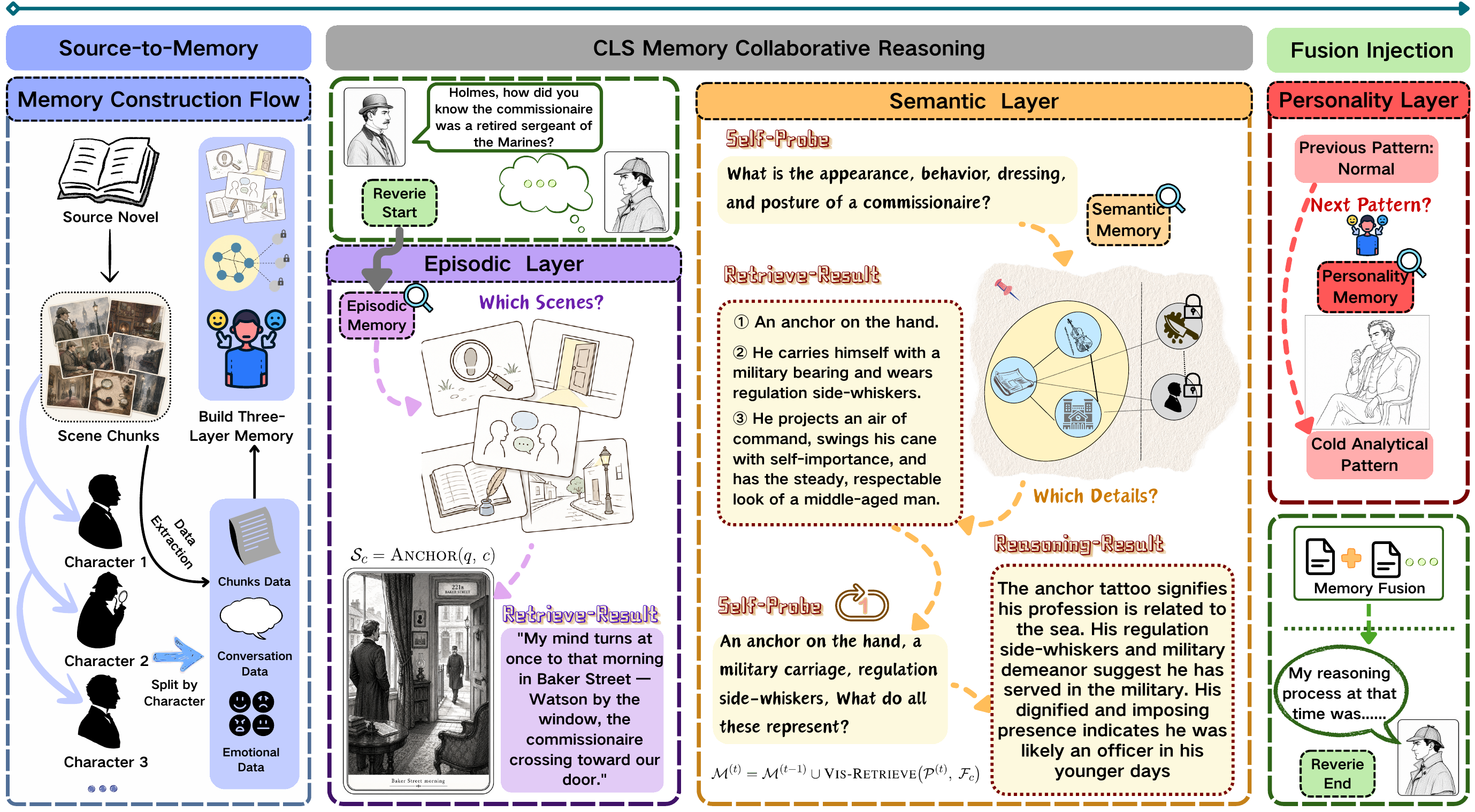}
   \caption{Overview of \ours. The framework comprises one offline construction phase and two inference phases for character $c$, as described in \S\ref{sec:method}: 1) \rmphase{rmBlue}{Source-to-Memory} extracts per-scene text, dialogue, and emotion data via an LLM and constructs a perspective-bounded three-layer memory for each focus character; 2) \rmphase{rmGray}{CLS Memory Collaborative Reasoning} anchors on the \rmlayer{rmViolet}{Episodic Layer} scene memory $\mathcal{S}_c$, which both grounds and informs \textsc{Self-Probe} to iteratively expand the fact pool $\mathcal{M}^{(t)}$ from the visibility-bounded \rmlayer{rmOrange}{Semantic Layer} subset $\mathcal{F}_c$, and finally synthesises a reasoning conclusion; 3) \rmphase{rmGreen}{Fusion Injection} selects a pattern $m^*$ from the \rmlayer{rmRed}{Personality Layer} and injects it into the final memory fusion that integrates memory components to respond from $c$'s perspective.}
   \label{fig:reveriemem}
\end{figure}
In this section we present \ours, a perspective-bounded three-layer memory architecture for character role-playing agents that addresses the failures identified in \S\ref{sec:intro}.
\ours therefore serves two design objectives, one for each failure:
\begingroup
\setlength\leftmargini{1em}
\begin{itemize}\setlength\itemsep{1pt}
   \item \textbf{Epistemic Objective.} The agent retrieves only facts the character could plausibly know in the source text, and abstains when its scene memories and retrieved facts do not support an answer.
   \item \textbf{Expressive Objective.} The agent's prose matches the style the character exhibits in comparable situations in the source text, rather than a flattened trait label.
\end{itemize}
\endgroup
\ours instantiates these objectives through three memory layers (\S\ref{sec:architecture}) and a perspective-bounded inference pipeline (\S\ref{sec:inference}).
Figure~\ref{fig:reveriemem} sketches the architecture.

\subsection{Overview}
\label{sec:overview}

\ours builds on two cognitive principles: the episodic--semantic distinction~\citep{cls1995} structures its memory layers (\S\ref{sec:architecture}), and reconstructive, scene-anchored recall~\citep{conway2000self} structures its inference pipeline (\S\ref{sec:inference}). Character knowledge is interpreted relative to the narrative point at which the agent responds, with per-character visibility assigned during memory construction. At inference, scene recall draws on the character's own episodic memories, while semantic retrieval is restricted to $\mathcal{F}_c$. The full system runs in three phases: an offline construction phase, \emph{Source-to-Memory} (\S\ref{sec:construction}), followed at inference by \emph{CLS Memory Collaborative Reasoning} and \emph{Fusion Injection} (both in \S\ref{sec:inference}).

\subsection{Bounded Three-Layer Memory}
\label{sec:architecture}

Following the episodic--semantic distinction (Figure~\ref{fig:reveriemem}), \ours splits factual memory into an \emph{episodic layer} (scene memories) and a \emph{semantic layer} (discrete facts with per-character visibility), and adds a \emph{personality layer} of discrete behavioural patterns derived from each character's conduct in the source text, selected at inference by emotion transitions~\citep{mcadams2013}.

\paragraph{Episodic Layer.}
\label{sec:episodic}
The episodic layer plays the hippocampus-like role of fast, scene-level encoding in CLS: it stores what the character has lived through as a perspectival frame that anchors subsequent retrieval. For each focus character $c$, the layer maintains a corpus of first-person scene summaries, one per scene $c$ was present for, recording in $c$'s voice what $c$ did, felt, perceived, and inferred about others. The corpus is indexed for similarity retrieval and serves as the substrate of the \textsc{Anchor} operation at inference time. Character-level scoping ensures every downstream retrieval starts from $c$'s lived experience.

\paragraph{Semantic Layer.}
\label{sec:semantic}
The semantic layer plays the neocortex-like role of slow, cross-episode integration: it consolidates the facts implied across scenes into a visibility-tagged knowledge graph. Each fact $f \in \mathcal{F}$ is a five-element \emph{SPOCV} tuple $f = (s, p, o, \kappa, V)$: $(s, p, o)$ is a standard subject--predicate--object triple; $\kappa$ records the in-narrative cause when one is explicit, capturing that narrative reasoning hinges on \emph{why} an event occurred, not only \emph{that} it did; and $V \subseteq \mathcal{C}$ is the per-character visibility set defined below.

Visibility is granted along four routes, distinguished by knowledge acquisition and sharing scope. Explicit, scene-grounded access comprises \emph{direct experience} through participation and \emph{observation} through non-participatory perception. For implicit access, we draw on common-ground theory~\citep{Clark_1996}: \emph{organisational sharing} covers information public within a relevant organisation, whereas \emph{common knowledge} covers general background shared across the narrative world. For each character $c$, semantic retrieval is restricted to $\mathcal{F}_c = \{f \in \mathcal{F} : c \in V(f)\}$. For \kbfqa, $V$ is assessed at each item's narrative point (\S\ref{sec:benchmark}). Appendix~\ref{sec:appendix-construction} provides assignment details; Table~\ref{tab:example-semantic} illustrates the four routes with source excerpts and recorded rationales.

\paragraph{Personality Layer.}
\label{sec:personality}
The personality layer addresses stylistic monotony: a character's behavioural style varies across situations such as confrontation, deference, and retreat. We therefore represent each character not as a static profile of trait adjectives but as a finite set of discrete patterns abstracted from the character's conduct in the source text.

For each character $c$, the layer maintains two artefacts. The first is a set $\{m_1, \ldots, m_K\}$ of \emph{personality patterns}, each pairing a short description with canonical excerpts from the source text (e.g.\ ``Lestrade in cautious deference to Holmes''); the descriptions support selection and the excerpts serve as in-context style anchors. The second is a record of $c$'s \emph{emotion transitions}, derived from per-utterance dialogue annotations: each transition encodes an $(\text{emotion}, \text{intensity}, \text{trigger})$ triple together with the speaker's inferred intent. At inference time, the emotion record situates next-state prediction within $c$'s observed trajectory, and the selected pattern provides the behavioural register that conditions memory fusion. The construction procedure is detailed in Appendix~\ref{sec:appendix-construction}.

\subsection{Source-to-Memory}
\label{sec:construction}
The three memory layers are constructed offline from the source novel by a scene-anchored procedure (Figure~\ref{fig:reveriemem}, left). The authors list focus characters with canonical names and aliases; a language model generates the remaining annotations.

The novel is partitioned into scenes in narrative order, and dialogues retain their original order within each scene. For each character, first-person scene summaries reflect only the character's local experience at the corresponding narrative point and are constructed without access to subsequent narrative material. Each knowledge tuple is labelled with its source scene ID.

Each SPOCV tuple records both its \texttt{visible\_to} assignment and a \texttt{visibility\_reasoning} field explaining the narrative evidence and knowledge-acquisition route. The authors verify the fact and its visibility rationale against the source scene, correcting or removing unsupported or ambiguous cases. Organisational sharing is retained only when the rationale supports that route and the information is public within the assigned organisational scope.

The detailed construction procedure and per-layer artefacts are provided in Appendix~\ref{sec:appendix-construction}, with per-book layer sizes in Appendix~\ref{sec:appendix-dataset}.

\subsection{Inference Pipeline}
\label{sec:inference}

\paragraph{CLS Memory Collaborative Reasoning.}
For a query $q$ directed at character $c$, the main reasoning chain first recalls relevant scenes and then reconstructs factual details.

\textit{Scene recall.} \textsc{Anchor} retrieves first-person scene summaries from $c$'s episodic corpus by similarity to $q$. For \kbfqa, retrieval is temporally restricted to memories available up to the item's narrative point; the question and options guide scene selection. Narrative continuation uses the current scene context. Both tasks dynamically select one or several relevant scenes from the responding character's memory:
\begin{equation}
\mathcal{S}_c = \textsc{Anchor}(q,\ c).
\end{equation}

\textit{Scene-guided detail reconstruction.} The scene memory $\mathcal{S}_c$ guides which factual details to recall from $\mathcal{F}_c$. Given $q$ and $\mathcal{S}_c$, \textsc{Self-Probe} generates a set of focused sub-queries $\mathcal{P}^{(1)}$ targeting the factual gaps in the scene memory, and the candidate-fact pool starts empty: $\mathcal{M}^{(0)} = \emptyset$. The reconstruction then loops over rounds $t = 1, \ldots, N$: each round issues the current probes against $\mathcal{F}_c$ via dense retrieval, expanding the pool
\begin{equation}
\mathcal{M}^{(t)} = \mathcal{M}^{(t-1)} \cup \textsc{Vis-Retrieve}\bigl(\mathcal{P}^{(t)},\ \mathcal{F}_c\bigr).
\end{equation}
The loop stops at round $\tau$ when the scene memory and retrieved facts support a response from $c$'s perspective, or when the round limit is reached. Otherwise, \textsc{Self-Probe} generates follow-up probes targeting the remaining gaps. After the loop, the CLS reasoning phase synthesises a conclusion $\rho_c$ from the recalled scenes $\mathcal{S}_c$ and retrieved facts $\mathcal{M}^{(\tau)}$ in response to $q$. The conclusion and its supporting memories are passed to memory fusion.

\paragraph{Fusion Injection.}
The personality track runs in parallel with the reasoning chain. When a salient emotional event occurs, an LLM selects a personality pattern $m^*$ based on the character's updated emotional state and dialogue context; otherwise, the previous pattern is retained.

\textit{Perspective-bounded memory fusion.} The closing memory-fusion stage integrates the reasoning conclusion $\rho_c$ with the terminal workspace $(\mathcal{S}_c, \mathcal{M}^{(\tau)})$, conditioned on the injected pattern $m^*$. The selected pattern's canonical excerpts and description serve as in-context style anchors, and \textsc{Memory-Fusion} produces a response from $c$'s perspective:
\begin{equation}
r = \textsc{Memory-Fusion}(q,\ c,\ \mathcal{S}_c,\ \mathcal{M}^{(\tau)},\ \rho_c;\ m^*).
\end{equation}
The response is grounded in $c$'s own scene memories and visibility-allowed facts. When these memories and facts do not support an answer, the prompt instructs the agent to state, in $c$'s own voice, that it lacks the information needed to answer. The factual memories constrain what the agent can claim, while the selected pattern guides how it responds. Algorithm~\ref{alg:reveriemem} (Appendix~\ref{sec:appendix-algorithm}) gives the procedure end-to-end, with selected inference prompt templates presented in Appendix~\ref{sec:appendix-prompts}.

\section{Knowledge Boundary Benchmark}
\label{sec:benchmark}

We construct \kbfqa, a multiple-choice benchmark for Knowledge Boundary Fidelity (KBF). It tests whether a character agent answers questions within its knowledge boundary at a given narrative point and refuses queries beyond that boundary. The dataset spans eight novels and contains 4{,}386 multiple-choice questions in total (2{,}442 \krf questions and 1{,}944 \kr questions). Each question is posed to a character $c$ and offers five options: four candidate answers $\{A, B, C, D\}$ and a refusal option $E$ (``I cannot answer this from my own knowledge''). Questions are extracted scene by scene and verified by the authors against their source scenes. Retrieval is restricted to episodic memories and semantic facts already available to the character by the item's narrative point. The visibility component $V$ (\S\ref{sec:semantic}) is assessed at each item's narrative point and determines the split: \krf items have the queried fact in $\mathcal{F}_c$ with gold answer in $\{A, B, C, D\}$; \kr items have it in $\mathcal{F} \setminus \mathcal{F}_c$ with gold answer $E$. The two splits target complementary epistemic skills: recalling what the character could know, and refusing what it could not. The \kr split tests character-specific access beyond factual correctness: an LLM that ``knows'' the fact still fails by committing to it while speaking as $c$. The construction process, example items, and per-book composition are in Appendix~\ref{sec:appendix-benchmark}.

\paragraph{The KBF metric.}
KBF summarises performance on the two KBF-QA splits. Let $\krf_{\text{acc}}, \kr_{\text{acc}}$ be the recall and refusal accuracies, with sample counts $n_{\krf}, n_{\kr}$. We define
\begin{equation}
   \kbf = \frac{n_{\krf} + n_{\kr}}{\dfrac{n_{\krf}}{\krf_{\text{acc}}} + \dfrac{n_{\kr}}{\kr_{\text{acc}}}},
   \label{eq:kbf}
\end{equation}
the sample-weighted harmonic mean. We define \kbf as zero when either split accuracy is zero. The harmonic mean (as in F1) penalises systems that collapse one split to gain the other; sample-weighting accounts for per-book variation in \krf:\kr ratio. Each method's free-form response is deterministically matched to one of $\{A, B, C, D, E\}$. Accuracy is computed against the item's gold option: the factual answer for \krf items and the refusal option $E$ for \kr items.

\section{Experiments}
\label{sec:experiments}

We evaluate \ours on two tasks over the same eight novels: \kbfqa assesses character knowledge boundaries, and pairwise narrative comparisons assess in-character generation.

\subsection{Evaluation Metrics}
\label{sec:metrics}

\paragraph{Boundary task.}
On KBF-QA, we report \krf, \kr, and \kbf as defined in \S\ref{sec:benchmark}.

\paragraph{Narrative-generation task.}
We follow the five-dimension pairwise evaluation protocol of \bookworld. An LLM judge compares two narratives generated for the same narrative instance and character cast. Four dimensions are shared across both modes: \textbf{Anthropomorphism} (An), \textbf{Character Fidelity} (CF), \textbf{Immersion \& Setting} (IS), and \textbf{Writing Quality} (WQ). The fifth dimension depends on the generation mode: \textbf{Storyline Quality} (SQ) for script-guided generation (\emph{with script}) and \textbf{Creativity} (Cr) for open-ended generation (\emph{without script}). For the narrative ablations, we also report CharacterBox scores~\citep{wang-etal-2025-characterbox} on seven dimensions (1--5 scale) and their mean (Appendix~\ref{sec:appendix-characterbox}).

\subsection{Experimental Setup}
\label{sec:setup}

\paragraph{Boundary task.}
All methods are evaluated on the benchmark in \S\ref{sec:benchmark} using the same \texttt{gpt-5-mini} backbone. \ours is compared against six reference methods: \emph{direct}, character-profile prompting without retrieval; four RAG baselines, \emph{naive RAG}~\citep{lewis-etal-2020-rag}, \emph{RAPTOR}~\citep{sarthi2024raptor}, \emph{HippoRAG}~\citep{hipporag}, and \emph{ComoRAG}~\citep{comorag2026}; and \emph{\bookworld}~\citep{bookworld}, a retrieval-augmented character-agent system.

\paragraph{Narrative-generation task.}
For each of the eight novels we adopt the scripts released by \emph{\bookworld} as experimental presets, yielding 226 scripts in total. \ours is evaluated with five generator LLMs: \texttt{gpt-5-mini}, \texttt{gemini-3.1}, \texttt{qwen-3.5-plus}, \texttt{deepseek-3.7}, and \texttt{deepseek-v4pro}~\citep{gpt-5-mini,qwen3.5,deepseek3.7}, two closed-source models and three open-source models. Reference methods are \emph{direct} (character-profile prompting), \emph{HoLLMwood}~\citep{hollmwood}, and \emph{\bookworld}. Pairwise judging uses \texttt{gpt-4o}. All reported win rates are directly measured from head-to-head comparisons of paired outputs on the same narrative instances under the same judging protocol. This protocol follows \citet{bookworld}, who reported substantial agreement between model-based pairwise judgments and human evaluations using Cohen's kappa coefficient ($\kappa$). We also evaluate the script-guided layer ablations using CharacterBox's criteria (Appendix~\ref{sec:appendix-characterbox}).

\begin{table}[t]
\centering
\footnotesize
\setlength{\tabcolsep}{4pt}
\renewcommand{\arraystretch}{1.0}
\begin{tabular}{@{}lccc@{}}
\toprule
\textbf{Method / Variant} & \textbf{\krf\,acc} & \textbf{\kr\,acc} & \textbf{\kbf} \\
\midrule
direct (no retrieval) & 36.4          & 38.2          & 37.2 \\
naive RAG             & 63.8          & 10.0          & 18.9 \\
RAPTOR                & 66.4          & 8.7           & 16.8 \\
HippoRAG              & 65.4          & 11.3          & 20.9 \\
ComoRAG               & 33.0          & 45.9          & 37.7 \\
\bookworld            & 41.6          & 35.5          & 38.7 \\
\ours                 & \textbf{68.1} & \textbf{81.2} & \textbf{73.3} \\
\midrule
w/o Episodic Layer           & 79.6 & 47.0 & 60.9 \\
w/o Semantic Layer           & 10.8 & 93.1 & 17.8 \\
\midrule
w/o Visibility Constraint    & 55.0 & 49.8 & 52.6 \\
Randomized Visibility Labels & 30.5 & 77.2 & 41.6 \\
\bottomrule
\end{tabular}
\caption{Boundary results (\%). Upper block: \ours and six reference methods, with the best scores in bold. Lower block: ablations of \ours. The two visibility variants retain the same Semantic Layer facts as the full system.}
\label{tab:task1-pooled}
\label{tab:ablation-task1}
\end{table}

\begin{table}[t]
\centering
\footnotesize
\setlength{\tabcolsep}{3pt}
\renewcommand{\arraystretch}{1.0}
\begin{tabular}{llccccc|ccccc}
\toprule
\multirow{2}{*}{\textbf{Model}} & \multirow{2}{*}{\textbf{Method}}
 & \multicolumn{5}{c|}{\textit{with script}}
 & \multicolumn{5}{c}{\textit{without script}} \\
 & & \textbf{An.} & \textbf{CF.} & \textbf{IS.} & \textbf{WQ.} & \textbf{SQ.}
   & \textbf{An.} & \textbf{CF.} & \textbf{IS.} & \textbf{WQ.} & \textbf{Cr.} \\
\midrule
\rowcolor{gray!15}
\multicolumn{12}{c}{\textit{closed-source models}} \\
\multirow{3}{*}{\texttt{gpt-5-mini}}
 & RM vs Direct & 94.3 & 92.8 & 100.0 & 100.0 & 90.9 & 81.4 & 97.3 & 95.9 & 97.7 & 95.0 \\
 & \hphantom{RM }vs HW & 66.0 & 83.7 & 92.2 & 100.0 & 99.6 & 67.7 & 69.2 & 88.5 & 96.2 & 60.0 \\
 & \hphantom{RM }vs BW & 68.3 & 93.1 & 67.7 & 90.5 & 96.3 & 65.3 & 67.6 & 78.2 & 94.4 & 92.6 \\
\midrule
\multirow{3}{*}{\texttt{gemini-3.1}}
 & RM vs Direct & 84.5 & 64.5 & 96.0 & 99.6 & 84.1 & 80.5 & 71.7 & 97.8 & 95.1 & 77.4 \\
 & \hphantom{RM }vs HW & 79.5 & 85.7 & 95.3 & 98.3 & 100.0 & 78.2 & 84.2 & 92.7 & 98.1 & 73.1 \\
 & \hphantom{RM }vs BW & 74.8 & 82.7 & 68.1 & 85.4 & 95.1 & 67.7 & 62.8 & 70.4  & 92.9  & 96.4 \\
\rowcolor{gray!15}
\multicolumn{12}{c}{\textit{open-source models}} \\
\multirow{3}{*}{\texttt{qwen-3.5-plus}}
 & RM vs Direct & 80.0 & 66.2 & 95.0 & 100.0 & 61.4 & 80.6 & 64.2 & 60.2 & 96.0 & 58.6 \\
 & \hphantom{RM }vs HW & 81.5 & 90.9 & 97.4 & 98.1 & 100.0 & 70.8 & 64.2 & 96.8 & 99.0 & 91.7 \\
 & \hphantom{RM }vs BW & 63.0 & 82.6 & 71.2 & 87.2 & 90.0 & 64.1 & 61.4 & 72.7  & 89.5 & 97.3 \\
\midrule
\multirow{3}{*}{\texttt{deepseek-3.7}}
 & RM vs Direct & 89.9 & 70.5 & 99.1 & 100.0 & 95.9 & 78.6 & 70.0 & 97.3 & 95.5 & 81.8 \\
 & \hphantom{RM }vs HW & 69.2 & 92.3 & 95.6 & 97.2 & 98.5 & 63.8 & 96.2 & 92.3 & 96.2 & 65.4 \\
 & \hphantom{RM }vs BW & 72.0 & 90.7 & 72.0 & 80.8 & 94.9 & 79.2 & 68.5 & 72.2  & 94.9 & 90.2 \\
\midrule
\multirow{3}{*}{\texttt{deepseek-v4pro}}
 & RM vs Direct & 38.3 & 57.4 & 73.9 & 98.3 & 53.2 & 33.9 & 30.4 & 62.6 & 80.9 & 92.2 \\
 & \hphantom{RM }vs HW & 42.0 & 79.1 & 84.3 & 98.3 & 96.5 & 52.2 & 60.0 & 87.0 & 96.5 & 78.3 \\
 & \hphantom{RM }vs BW & 56.5 & 73.0 & 70.4 & 91.3 & 84.3 & 60.0 & 50.4 & 80.0 & 96.5 & 75.7 \\
\bottomrule
\end{tabular}
\caption{Direct head-to-head win rates (\%) of \ours against Direct, HoLLMwood (HW), and \bookworld (BW) under two narrative-generation modes. Protocol: Appendix~\ref{sec:appendix-task2}.}
\label{tab:task2-full}
\end{table}

\subsection{Main Results}
\label{sec:main-results}

\paragraph{Knowledge-boundary fidelity.}
The upper block of Table~\ref{tab:task1-pooled} compares \ours with six reference methods. \ours achieves higher \krf, \kr, and \kbf scores than all six methods. The RAG baselines show different recall--refusal profiles: \emph{naive RAG}, \emph{RAPTOR}, and \emph{HippoRAG} achieve moderate recall accuracy but low refusal accuracy; \emph{ComoRAG} has higher refusal but lower recall accuracy than these three baselines.

\paragraph{In-character narrative generation.}
Table~\ref{tab:task2-full} reports direct head-to-head win rates of \ours against each reference method across five generator LLMs. For the first four generator LLMs, \ours exceeds 50\% on every dimension against all three reference methods under both modes, with particularly strong gains on WQ. The results for \texttt{deepseek-v4pro} are more mixed, especially on An and CF against \emph{Direct}, indicating that the gains depend on the backbone and evaluation dimension. Against \bookworld, \ours wins a majority of comparisons on every dimension for all five generator LLMs under both modes.

\subsection{Ablation Study}
\label{sec:ablation}

\paragraph{Boundary ablation.}
The lower block of Table~\ref{tab:ablation-task1} reports ablations of the factual memory layers and the visibility mechanism. Removing the episodic layer increases recall but reduces refusal, whereas removing the semantic layer sharply lowers recall and increases refusal. Both variants underperform the full system on \kbf, showing that scene memories and structured facts contribute differently to balancing factual recall and refusal.

To isolate the effect of visibility, we further retain the same Semantic Layer facts and either remove the visibility constraint or randomize its labels. Removing the constraint reduces \kbf from 73.3 to 52.6 (20.7 percentage points), with refusal accuracy falling from 81.2 to 49.8. Randomizing visibility labels reduces \kbf to 41.6 (31.7 percentage points), with recall accuracy falling from 68.1 to 30.5. Both visibility variants underperform the full system on every book. With semantic facts held fixed, these results show that accurate character--fact visibility supports both factual recall and boundary-respecting refusal.

\paragraph{Narrative ablation.}
Table~\ref{tab:ablation-task2} reports the full system's pairwise win rate against each ablated variant under both modes. The full system is preferred on most dimensions, with the highest win rates on WQ across all ablations. Against variants without either factual layer, the full system achieves IS win rates of 68.3--83.3\%, supporting the contribution of scene-anchored knowledge. Against the variant without the personality layer, the full system has higher win rates on IS and WQ than on CF in both generation modes. CF emphasises consistency with a character's established traits and background. Without the personality layer, \bookworld's fixed preset lines and character behaviours remain available. In open-ended generation, these cues may sustain character consistency even when expression varies little across situations.

\paragraph{CharacterBox evaluation.}
To complement CF with assessments of emotional expression and personality consistency, we evaluate \ours and its layer ablations using CharacterBox's criteria in script-guided generation (Appendix~\ref{sec:appendix-characterbox}). The full system outperforms all three layer ablations on six of seven dimensions, averaging 4.70 versus 3.35--3.41. Removing the personality layer reduces Emotional Expression (EE) from 5.00 to 3.03 and Personality Traits (PT) from 4.93 to 3.30, while knowledge accuracy remains similar (4.88 vs.\ 4.80), supporting the layer's contribution to character-grounded expression. The full system's lower Adaptability (AD) score (3.22 vs.\ 3.37--3.47) suggests a consistency--adaptability trade-off.

\begin{table}[t]
\centering
\footnotesize
\setlength{\tabcolsep}{3pt}
\renewcommand{\arraystretch}{1.0}
\begin{tabular}{lccccc}
\toprule
\textbf{Comparison} & \textbf{An.} & \textbf{CF.} & \textbf{IS.} & \textbf{WQ.} & \textbf{SQ./Cr.} \\
\midrule
\rowcolor{gray!15}
\multicolumn{6}{c}{\textit{with script}} \\
w vs w/o Episodic    & 61.7 & 66.7 & 83.3 & 86.7 & 75.0 \\
w vs w/o Semantic    & 66.7 & 65.0 & 81.7 & 86.7 & 81.7 \\
w vs w/o Personality & 61.7 & 58.3 & 66.7 & 93.3 & 86.7 \\
\rowcolor{gray!15}
\multicolumn{6}{c}{\textit{without script}} \\
w vs w/o Episodic    & 53.3 & 51.7 & 68.3 & 91.7 & 76.7 \\
w vs w/o Semantic    & 51.7 & 53.3 & 80.0 & 96.7 & 75.0 \\
w vs w/o Personality & 55.0 & 50.0 & 75.0 & 93.3 & 85.0 \\
\bottomrule
\end{tabular}
\caption{Pairwise win rates (\%) of \ours against ablated variants that remove one memory layer at a time.}
\label{tab:ablation-task2}
\end{table}

\subsection{Discussion}
\label{sec:discussion}

The two tasks support coordinating factual access and situated expression within one memory architecture. On \kbfqa, controlled ablations show that accurate character--fact visibility improves recall and refusal with semantic facts held fixed. In narrative generation, the layer ablations and CharacterBox scores support the complementary roles of scene-grounded knowledge and situation-dependent behavioural patterns.

\paragraph{Backbone dependence.}
The contrast between \texttt{deepseek-3.7} and \texttt{deepseek-v4pro} in Table~\ref{tab:task2-full} suggests an interaction between the backbone and the memory pipeline. Differences in model scale, attention architecture, and post-training~\citep{DBLP:journals/corr/abs-2606-19348} may affect native reasoning, the use of retrieved memories, and fine-grained style control. \bookworld also reports that its Llama-3.3-70B pipeline underperforms simpler baselines on several dimensions~\citep{bookworld}. The results motivate studying backbone--memory interactions.

\paragraph{Inference cost.}
The three memory layers are constructed once offline and reused across interactions. Under the same backbone and execution setting, \ours uses 7.4\% more tokens and takes 11.6\% longer per character response than Single-pass (Appendix~\ref{sec:appendix-cost}).

\section{Conclusion}
\label{sec:conclusion}

We propose \ours, a three-layer memory architecture inspired by cognitive psychology, to address Factual Overreach and Stylistic Monotony in role-playing agents. It anchors reasoning in characters' own experiences, limits retrieval to facts they could know, and guides expression through situation-dependent behavioural patterns. Experiments across eight novels using our \kbfqa benchmark and narrative evaluation show gains over the evaluated baselines. These findings support grounding knowledge and expression in memory organised around each character's perspective.

\section*{Limitations}
\label{sec:limitations}
\kbfqa targets the cognitive side of perspective: what a character could plausibly know given their narrative position. Targeted stress tests of \emph{Rashomon}-style multi-witness contradictions, deliberately information-asymmetric detective plots, and unreliable narrators remain future work.

\paragraph{Multi-character interaction.}
Within multi-character simulations, \ours focuses on each character's memory and response generation. Explicitly modelling how characters reason about one another's knowledge and inferences remains a direction for future work.

\paragraph{Benchmark validation.}
We verify \kbfqa items and memory visibility assignments against their source scenes. Extending this validation with independent re-annotation and inter-annotator agreement analysis remains future work.

\section*{Ethics Statement}
\label{sec:ethics}

In this paper, we introduce \ours, a perspective-bounded memory architecture for character role-playing agents, together with the \kbfqa benchmark. The development and use of both are guided by ethical principles to ensure responsible and beneficial outcomes.

\paragraph{Data sources.}
Our experiments cover eight novels: six public-domain works in English (\emph{A Study in Scarlet}, \emph{The Adventures of Tom Sawyer}, \emph{Treasure Island}, \emph{Around the World in Eighty Days}, \emph{Dracula}, \emph{Paradise Lost}) and two works in Chinese (\emph{Solaris}, in its Chinese translation, and \emph{Ball Lightning}). We use these texts strictly for academic and non-commercial research purposes; no full text of any work is redistributed with the paper, and all references in the benchmark are presented as short, scene-anchored excerpts.

\paragraph{Use of Human Annotations.}
\kbfqa is constructed jointly by an LLM and human annotators who have read each of the eight novels in their entirety. The annotators are the authors of this work; no external workers were recruited and no compensation was distributed. The annotation task is restricted to two operations: (i) verifying each LLM-drafted item's factual grounding and character visibility against the source text, and (ii) discarding items whose visibility assignment is borderline or ambiguous. No personal data, demographic information, or third-party content beyond the eight novels is collected, stored, or released at any stage. The annotators were informed of the purpose of the verification task and of how their annotations would be used in the released benchmark.

\paragraph{Risk.}
\ours inherits the general risks of LLM-based systems: hallucination, bias propagated from pretraining data, and the possibility of generating misleading or harmful content. Our perspective-bounded design directly mitigates one specific failure mode --- factual overreach across character knowledge boundaries --- but does not address these broader base-model risks. Like other character-agent systems, \ours could in principle be misused to produce content that misrepresents a real person; we intend it to be used only with virtual characters, and we discourage any application to the simulation of real, identifiable individuals. We encourage the responsible use of \ours for academic, educational, and creative purposes, and discourage any application aimed at deception, harassment, or impersonation.
\paragraph{Use of AI Assistants.}
AI writing assistants (large language models) were used for grammar polishing, sentence-level rewording, code-level debugging support, and bibliographic source lookup and verification. The human authors are responsible for the research design, experimental decisions, analysis, and final manuscript. AI writing assistants were not used to generate experimental results.

\bibliography{custom}

\clearpage
\appendix
\makeatletter
\setlength{\@fptop}{0pt}
\setlength{\@fpsep}{14pt plus 2pt}
\makeatother

\section{Three-Layer Memory Statistics}
\label{sec:appendix-dataset}

We build the three memory layers for eight novels (six English, two Chinese), covering 50 focus characters across 360 scenes. The Semantic Layer (\S\ref{sec:semantic}) holds 12{,}468 SPOCV tuples and the Personality Layer (\S\ref{sec:personality}) holds 1{,}536 behavioural patterns; per-book breakdowns are in Table~\ref{tab:memory-stats}. Per-book \kbfqa{} composition and scores are reported in Appendix~\ref{sec:appendix-benchmark}.

\section{Source-to-Memory Construction and Examples}
\label{sec:appendix-construction}

\paragraph{Scene preparation.}
The novel is partitioned into scenes by an LLM-guided procedure that merges contiguous text spans falling within a single dramatic unit. Scenes retain their order in the source text, and dialogues retain their original order within each scene. Each scene is augmented with situational metadata (location, time, atmosphere) and a roster indicating whether each focus character is present and active, present but silent, or only referenced. These scenes provide a common basis for all three memory layers.

\paragraph{Episodic summarisation.}
For each focus character $c$ and each scene in which $c$ is present, we elicit a first-person retrospective summary in $c$'s voice. Summaries are constructed scene by scene, without access to subsequent narrative material, so that each summary reflects only the character's local experience at that point. The resulting per-character corpora are indexed using dense embeddings for similarity-based retrieval at inference time.

\paragraph{Knowledge graph extraction.}
Following an extract-then-relate procedure~\citep{hipporag, comorag2026}, we first extract entities from each scene, then induce $(s, p, o, \kappa)$ relations over them, with $\kappa$ recording the in-narrative cause when explicit. Entity aliases are unified across scenes, after which facts are deduplicated against the canonical entity table. Each knowledge tuple is labelled with its source scene ID. The visibility component $V$ is then assigned in two phases, completing the SPOCV tuple. A local phase assigns direct experience to participants, observation to non-participating witnesses, and common knowledge for world-level background. A propagation phase extends organisational sharing through a character--organisation membership graph derived from the same scenes. Source-scene verification checks the sharing route and whether the information is public within the assigned scope (\S\ref{sec:construction}).

\paragraph{Personality pattern construction.}
We extract the dialogues attributed to each focus character on a per-scene basis and normalise speaker labels to canonical names. Each utterance is annotated with an $(\text{emotion}, \text{intensity}, \text{trigger})$ triple together with the speaker's inferred intent; concatenated in narrative order, these annotations form the character's emotion-transition record. We derive $K$ situation-dependent patterns from the character's speech and behaviour across scenes. Each pattern pairs a short description used for selection with canonical dialogue excerpts that serve as style anchors during memory fusion.

\paragraph{Memory examples.}

Tables~\ref{tab:example-episodic}--\ref{tab:example-personality} present artefacts produced by our Source-to-Memory pipeline (\S\ref{sec:construction}). The episodic example comes from Scene~1 of \emph{A Study in Scarlet}, and the personality examples from Scenes~1 and~25. Table~\ref{tab:example-semantic} illustrates the four visibility routes with stored facts from this novel, \emph{Around the World in Eighty Days}, and \emph{Tom Sawyer}. Each fact is paired with a source excerpt and its verbatim \texttt{visibility\_reasoning}. The examples contrast participation with observation of the same event and illustrate both information publicly available within an organisation and seasonal background knowledge shared in the story's setting.

\section{Inference Details and Algorithm}
\label{sec:appendix-algorithm}

\paragraph{Inference overview.}
\ours's inference pipeline realises a perspective-bounded form of reconstructive recall. Given a query $q$ directed at character $c$, \emph{CLS Memory Collaborative Reasoning} recalls relevant scenes from $c$'s episodic memory and iteratively retrieves facts from the visibility-allowed subset $\mathcal{F}_c$. \emph{Fusion Injection} integrates the resulting reasoning conclusion and supporting memories, conditioned on the pattern $m^*$ supplied by the parallel personality track (Figure~\ref{fig:reveriemem}, middle and right).

\paragraph{Scene-guided reasoning.}
The recalled scene memory $\mathcal{S}_c$ provides a character-specific frame for factual detail reconstruction, following the Self-Memory System's hierarchical retrieval cascade~\citep{conway2000self}. \textsc{Self-Probe} uses this memory to identify factual gaps, and \textsc{Vis-Retrieve} searches the visibility-allowed subset $\mathcal{F}_c$ to expand the candidate pool. Follow-up probes draw on the recalled scenes and accumulated facts to address the remaining gaps.

\paragraph{Personality pattern selection.}
The personality track runs in parallel with the reasoning chain and supplies a pattern $m^*$ to the final memory-fusion stage through three operations: \textsc{Emo-Gate}, \textsc{Emo-Transition}, and \textsc{Pattern}.

\textsc{Emo-Gate} judges whether the current dialogue turn carries a salient emotional event; routine exchanges leave the emotional state in place. When the gate fires, \textsc{Emo-Transition} reads $c$'s emotion-transition record and proposes a new state, and \textsc{Pattern} selects $m^*$ from $\{m_1, \ldots, m_K\}$ using an LLM conditioned on the new emotional state and dialogue context.

\paragraph{Full algorithm.}

Algorithm~\ref{alg:reveriemem} presents the full inference procedure (\S\ref{sec:inference}). Semantic retrieval is restricted to $\mathcal{F}_c$, and reasoning is anchored on $c$'s own scene memory $\mathcal{S}_c$ before factual detail is reconstructed. The scene-guided loop runs for at most $N$ rounds and exits early once the workspace $(\mathcal{S}_c, \mathcal{M}^{(t)})$ supports a response.

\section{\kbfqa Construction}
\label{sec:appendix-benchmark}

We build \kbfqa from LLM-drafted candidate items grounded in individual source scenes. The authors, who have read each novel, verify each item's factual grounding and character visibility against its source scene at the corresponding narrative point. Items with borderline or ambiguous visibility assignments are discarded. Before retrieval, candidate episodic memories and semantic facts are restricted to those available to the character at the item's narrative point, excluding knowledge introduced only in later scenes. Figure~\ref{fig:benchmark-examples} illustrates one example from each split (posed to Dr.\ John Seward), and Table~\ref{tab:task1-by-book} reports per-book item counts and \kbf scores for \ours and the six reference methods (\S\ref{sec:experiments}).

\section{Pairwise Narrative Evaluation Protocol}
\label{sec:appendix-task2}

We use the eight novels and their plot scripts from the released setup of \citet{bookworld}. Table~\ref{tab:task2-by-book} gives the per-book counts, totalling 226 scripts.

For each generator LLM and generation mode, we directly compare \ours with \emph{Direct}, \emph{HoLLMwood}, and \emph{\bookworld} by pairing outputs for the same narrative instance and applying the judging protocol of \citet{bookworld}. Table~\ref{tab:task2-full} reports the win rate (\%) for each pair and evaluation dimension. The mean win rate of 78.86\% against \bookworld is the unweighted average of the 50 entries across five generator LLMs, two generation modes, and five evaluation dimensions in Table~\ref{tab:task2-full}.

The layer ablations in Table~\ref{tab:ablation-task2} use \texttt{gpt-5-mini} as the agent backbone and \texttt{gpt-4o} for pairwise judging. They cover 60 narrative instances per generation mode across the eight novels, using the first quarter of each book's scripts (rounded up).

\begin{table}[!htbp]
\centering
\small
\setlength{\tabcolsep}{4pt}
\renewcommand{\arraystretch}{1.1}
\begin{tabularx}{\columnwidth}{@{}Xlc@{}}
\toprule
\textbf{Book} & \textbf{Language} & \textbf{\# Scripts} \\
\midrule
A Study in Scarlet           & English & 12 \\
Tom Sawyer                   & English & 36 \\
Treasure Island              & English & 34 \\
Around the World in 80 Days  & English & 37 \\
Dracula                      & English & 42 \\
Paradise Lost                & English & 25 \\
\midrule
Solaris                      & Chinese & 13 \\
Ball Lightning               & Chinese & 27 \\
\midrule
\textbf{Total}               & ---     & \textbf{226} \\
\bottomrule
\end{tabularx}
\caption{Per-book counts of plot scripts from \citet{bookworld} used for narrative generation.}
\label{tab:task2-by-book}
\end{table}

\section{Inference Prompts}
\label{sec:appendix-prompts}

Tables~\ref{tab:prompt-self-probe}--\ref{tab:prompt-memory-fusion} present prompt templates for \textsc{Self-Probe} (Algorithm~\ref{alg:reveriemem}, lines~2 and~8), \textsc{Pattern} (line~13), and \textsc{Memory-Fusion} (line~17). They respectively generate retrieval probes, select a behavioural pattern, and produce the final response from the character's perspective. \textsc{Anchor} and \textsc{Vis-Retrieve} are dense-retrieval operations.

The memory-fusion prompt draws factual support from the character's scene memories and retrieved facts, while the selected behavioural pattern guides expression. When these memories and facts do not support an answer, the prompt instructs the agent to state, in the character's own voice, that it lacks the information needed to answer.

\section{CharacterBox Evaluation}
\label{sec:appendix-characterbox}

We evaluate \ours and its layer ablations using CharacterBox's criteria~\citep[Appendix~G]{wang-etal-2025-characterbox} on all 226 script-guided instances from the eight novels, with \texttt{gpt-5-mini} as the agent backbone. For each instance, \texttt{gpt-4o} assigns each narrative in the full--ablation pair a score from 1 to 5 on each dimension.

The seven dimensions are Knowledge Accuracy (KA), Behavioural Accuracy (BA), Emotional Expression (EE), Personality Traits (PT), Immersion (IM), Adaptability (AD), and Behavioural Coherence (BC). EE assesses the richness and contextual appropriateness of emotional expression, while PT assesses consistency with the character's core personality traits.

Table~\ref{tab:characterbox-ablation} reports mean scores for each variant; AVG is the mean across the seven dimensions.

\section{Online Inference Cost}
\label{sec:appendix-cost}

The three memory layers are constructed once offline and reused across interactions. At inference, the Semantic Layer retrieves compact, visibility-allowed SPOCV facts rather than full source passages. Table~\ref{tab:online-cost} reports average online token usage and latency per character response under the same backbone and execution setting, excluding offline memory construction.

\begin{table}[!htbp]
\centering
\small
\setlength{\tabcolsep}{7pt}
\renewcommand{\arraystretch}{1.1}
\begin{tabular}{lrr}
\toprule
\textbf{Method} & \textbf{Total tokens} & \textbf{Time (s)} \\
\midrule
Direct & 1,655 & 2.96 \\
Single-pass & 3,613 & 5.77 \\
\ours & 3,881 & 6.44 \\
\bottomrule
\end{tabular}
\caption{Average online costs per character response.}
\label{tab:online-cost}
\end{table}

\begin{table}[!htbp]
\centering
\small
\setlength{\tabcolsep}{8pt}
\renewcommand{\arraystretch}{1.15}
\begin{tabular}{lccccc}
\toprule
\textbf{Book} & \textbf{Language} & \textbf{\# Characters} & \textbf{\# Scenes} & \textbf{\# SPOCV facts} & \textbf{\# Patterns} \\
\midrule
A Study in Scarlet           & English & 5  & 25 & 904     & 197 \\
Tom Sawyer                   & English & 3  & 92 & 2{,}298 & 170 \\
Treasure Island              & English & 3  & 46 & 1{,}517 & 126 \\
Around the World in 80 Days  & English & 10 & 41 & 1{,}836 & 175 \\
Dracula                      & English & 9  & 33 & 973     & 166 \\
Paradise Lost                & English & 7  & 38 & 1{,}926 & 167 \\
\midrule
Solaris                      & Chinese & 4  & 50 & 1{,}657 & 134 \\
Ball Lightning               & Chinese & 9  & 35 & 1{,}357 & 401 \\
\midrule
\textbf{Total}               & ---     & \textbf{50} & \textbf{360} & \textbf{12{,}468} & \textbf{1{,}536} \\
\bottomrule
\end{tabular}
\caption{Per-book corpus and memory statistics.}
\label{tab:memory-stats}
\end{table}

\begin{table}[!htbp]
\centering
\small

\noindent\begin{tcolorbox}[
  colback=rmViolet!12, colframe=rmViolet!12,
  arc=0pt, boxrule=0pt, sharp corners,
  left=4pt, right=4pt, top=2pt, bottom=2pt,
  width=\textwidth, before skip=0pt, after skip=0pt,
  halign=left,
]
{\color{rmViolet!75!black}\bfseries Episodic Layer \textnormal{\itshape\;· first-person scene summary for Dr.\ Watson, Scene~1}}
\end{tcolorbox}
\renewcommand{\arraystretch}{1.25}
\arrayrulecolor{rmViolet!70!black}
\begin{tabularx}{\textwidth}{@{}p{2.6cm}X@{}}
\specialrule{0.6pt}{0pt}{2pt}
\textbf{Scene} & 1 --- Chemical laboratory at St.\ Bartholomew's Hospital, London \\
\addlinespace[2pt]
\textbf{Atmosphere} & Intellectually intense, slightly eccentric; contrasts Watson's weariness with Holmes's energetic obsession \\
\addlinespace[2pt]
\textbf{Scene understanding} & \textit{``I had recently returned to England after being wounded during the Second Anglo-Afghan War, and was seeking affordable lodgings. By chance I met Stamford at the Criterion Bar; over lunch he mentioned a man named Sherlock Holmes who was also looking for someone to share rooms\ldots\ At Bart's, Holmes was ecstatic over a reagent that detects haemoglobin, and demonstrated it by pricking his own finger. Stamford introduced us, and Holmes startled me by correctly deducing I had been in Afghanistan.''} \\
\addlinespace[2pt]
\textbf{My actions} & \textit{``I asked Stamford what he'd been doing; pressed him on Holmes's character when he seemed evasive; agreed to visit the lab; expressed polite interest in the demonstration; was taken aback by Holmes's abrupt observation about Afghanistan.''} \\
\addlinespace[2pt]
\textbf{My emotions} & \textit{``Initial surprise and warmth at a familiar face; cautious optimism about the lodgings; mild suspicion at Stamford's hesitation; strong surprise at Holmes's Afghanistan deduction; restrained admiration mixed with unease.''} \\
\specialrule{0.8pt}{2pt}{0pt}
\end{tabularx}
\arrayrulecolor{black}

\caption{Episodic Layer artefact: Watson's first-person summary of Scene~1 of \emph{A Study in Scarlet}.}
\label{tab:example-episodic}
\end{table}

\begin{table}[!tp]
\centering
\small
\setlength{\tabcolsep}{4pt}
\renewcommand{\arraystretch}{1.13}
\tcbset{visibilityexample/.style={
  colback=rmOrange!3, colframe=rmOrange!55!black,
  colbacktitle=rmOrange!12, coltitle=rmOrange!65!black,
  fonttitle=\bfseries, boxrule=0.4pt, arc=1pt,
  left=6pt, right=6pt, top=4pt, bottom=4pt,
  before skip=0pt, after skip=6pt, width=\textwidth
}}

\begin{tcolorbox}[visibilityexample,title={(a--b) Direct experience and observation}]
\emph{A Study in Scarlet}, Scene~23: Hope recounts following Drebber and Stangerson.\par\smallskip
\begin{tabularx}{\linewidth}{@{}p{2.55cm}X@{}}
\textbf{Fact $(s,p,o)$} & (Drebber, went to, Euston Station) \\
\textbf{Visible to} & Jefferson Hope; Joseph Stangerson \\
\textbf{Source} & ``Drebber and Stangerson followed it, and drove off.'' \ldots\ ``At Euston Station they got out, and I left a boy to hold my horse, and followed them on to the platform.'' \\
\textbf{Rationale} & ``Public event witnessed by Jefferson and participated in by Stangerson.'' \\
\textbf{(a) Participation} & Stangerson travels with Drebber. \\
\textbf{(b) Observation} & Hope follows them and witnesses their arrival. \\
\end{tabularx}
\end{tcolorbox}

\begin{tcolorbox}[visibilityexample,title={(c) Organisational sharing}]
\emph{Around the World in Eighty Days}, Scene~6: Fogg's photograph is displayed at the club.\par\smallskip
\begin{tabularx}{\linewidth}{@{}p{2.55cm}X@{}}
\textbf{Fact $(s,p,o)$} & (Reform Club, displayed, photograph of Phileas Fogg) \\
\textbf{Shared scope} & Reform Club (stored \texttt{organization\_visibility}) \\
\textbf{Source} & ``His photograph, which was hung with those of the rest of the members at the Reform Club, was minutely examined'' \\
\textbf{Rationale} & ``Club members and visitors can see the photo; relevant to the organization.'' \\
\textbf{Access basis} & The displayed photograph is available within the club. \\
\end{tabularx}
\end{tcolorbox}

\begin{tcolorbox}[visibilityexample,title={(d) Common knowledge}]
\emph{Tom Sawyer}, Scene~19: summer evenings in St.~Petersburg.\par\smallskip
\begin{tabularx}{\linewidth}{@{}p{2.55cm}X@{}}
\textbf{Fact $(s,p,o)$} & (summer evenings, are, long) \\
\textbf{Visible to} & Huckleberry Finn; Thomas Sawyer; Joe Harper \\
\textbf{Source} & ``The summer evenings were long. It was not dark, yet.'' \\
\textbf{Rationale} & ``Universal seasonal fact; common sense knowledge.'' \\
\textbf{Access basis} & Seasonal background knowledge shared within the story's setting. \\
\end{tabularx}
\end{tcolorbox}

\caption{Semantic Layer examples illustrating the four visibility routes. Facts and selected access fields are shown with source excerpts and verbatim \texttt{visibility\_reasoning} (``Rationale''). Route annotations follow the definitions in \S\ref{sec:semantic}.}
\label{tab:example-semantic}
\end{table}

\begin{table}[!tp]
\centering
\small

\noindent\begin{tcolorbox}[
  colback=rmRed!12, colframe=rmRed!12,
  arc=0pt, boxrule=0pt, sharp corners,
  left=4pt, right=4pt, top=2pt, bottom=2pt,
  width=\textwidth, before skip=0pt, after skip=0pt,
  halign=left,
]
{\color{rmRed!75!black}\bfseries Personality Layer \textnormal{\itshape\;· two of fifty personality patterns for Sherlock Holmes}}
\end{tcolorbox}
\renewcommand{\arraystretch}{1.3}
\arrayrulecolor{rmRed!70!black}
\begin{tabularx}{\textwidth}{@{}llXX@{}}
\specialrule{0.6pt}{0pt}{2pt}
\textbf{Pattern} & \textbf{Scene} & \textbf{Description} & \textbf{Canonical excerpt} \\
\arrayrulecolor{black}\midrule
$m_{1}$  & 1  & \textit{Escalating, unrestrained joy --- shouting, running, chuckling; voice bright and urgent, gestures expansive and physically energetic; tone exuberant and self-absorbed.} & ``I've found it! I've found it!'' \\
\addlinespace[3pt]
$m_{49}$ & 25 & \textit{Calm detachment; flat intonation and minimal physical emphasis; grants permission and redirects attention without warmth or irritation.} & ``You may do what you like, Doctor.'' \\
\arrayrulecolor{rmRed!70!black}\specialrule{0.8pt}{2pt}{0pt}
\end{tabularx}
\arrayrulecolor{black}

\caption{Personality Layer artefacts: two of fifty patterns for Sherlock Holmes, drawn from opposite ends of \emph{A Study in Scarlet}.}
\label{tab:example-personality}
\end{table}

\begin{algorithm}[!tp]
\small
\begin{algorithmic}[1]
\State $\mathcal{S}_c \gets \textsc{Anchor}(q,\ c)$ \Comment{scene recall}
\State $\mathcal{P}^{(1)} \gets \textsc{Self-Probe}(q,\ \mathcal{S}_c)$;\quad $\mathcal{M}^{(0)} \gets \emptyset$
\For{$t = 1$ \textbf{to} $N$} \Comment{scene-guided detail reconstruction}
    \State $\mathcal{M}^{(t)} \gets \mathcal{M}^{(t-1)} \cup \textsc{Vis-Retrieve}(\mathcal{P}^{(t)},\ \mathcal{F}_c)$
    \State $\sigma^{(t)} \gets \textsc{Sufficient}(q,\ \mathcal{S}_c,\ \mathcal{M}^{(t)})$
    \If{$\sigma^{(t)} = \text{true}$ \textbf{or} $t = N$} \textbf{break} \EndIf
    \State $\mathcal{P}^{(t+1)} \gets \textsc{Self-Probe}(q,\ \mathcal{S}_c,\ \mathcal{P}^{(1:t)},\ \mathcal{M}^{(t)})$
\EndFor
\State $\tau \gets t$ \Comment{exit round}
\State $\rho_c \gets$ reasoning conclusion for $q$ grounded in $(\mathcal{S}_c,\mathcal{M}^{(\tau)})$
\If{$\textsc{Emo-Gate}(q,\ c) = \textsc{Fire}$} \Comment{parallel personality track}
    \State $e^* \gets \textsc{Emo-Transition}(c)$;\quad $m^* \gets \textsc{Pattern}(c,\ e^*)$
\Else
    \State $m^* \gets$ previous pattern of $c$ \Comment{emotional inertia}
\EndIf
\State \Return $r \gets \textsc{Memory-Fusion}(q,\ c,\ \mathcal{S}_c,\ \mathcal{M}^{(\tau)},\ \rho_c;\ m^*)$ \Comment{perspective-bounded memory fusion}
\end{algorithmic}
\caption{\ours inference for query $q$ directed at character $c$.}
\label{alg:reveriemem}
\end{algorithm}

\begin{figure}[!tp]
\centering
\begin{minipage}{0.97\textwidth}
\small

\textit{Source:} \textcolor{rmBlue}{\textit{Dracula}} \hfill
\textit{Asked of:} \textcolor{rmGreen!55!black}{Dr.\ John Seward}

\vspace{4pt}

\begin{tcolorbox}[
  enhanced,
  colback=rmGreen!4, colframe=rmGreen!70!black,
  arc=3pt, boxrule=0.6pt,
  title={\textbf{KRf}\,\textnormal{\itshape\,· fact visible to Seward\,\ ($f \in \mathcal{F}_{\text{Seward}}$)}},
  coltitle=white, colbacktitle=rmGreen!70!black,
  fonttitle=\bfseries,
  left=8pt, right=8pt, top=5pt, bottom=5pt,
]
\textit{Inquirer:} \textit{``Dr.\ Seward, you saw your patient slip away one night --- where did he go?''}

\smallskip
\textit{Options:}\\
\quad(A) the ruined churchyard\\
\quad(B) the quay at Whitby\\
\quad(C) a deserted house\\
\quad(D) an empty lodging in Munich\\
\quad(E) ``I cannot answer this from my own knowledge.''

\smallskip
\textsc{Gold}~\textcolor{rmGreen!55!black}{\textbf{(C)}} \hfill
\textsc{\bookworld}~\textcolor{rmRed}{(A)}\,{\large\textcolor{rmRed}{$\times$}} \hfill
\textsc{\ours}~\textcolor{rmGreen!55!black}{(C)}\,{\large\textcolor{rmGreen!55!black}{$\checkmark$}}

\smallskip
\textit{Why:} After Renfield escapes, Seward and the attendants find him at the old chapel door in the grounds of the deserted house. Seward thus knows the destination from direct observation, making (C) the correct answer.
\end{tcolorbox}

\vspace{4pt}

\begin{tcolorbox}[
  enhanced,
  colback=rmOrange!4, colframe=rmOrange!70!black,
  arc=3pt, boxrule=0.6pt,
  title={\textbf{KR}\,\textnormal{\itshape\,· fact not visible to Seward\,\ ($f \in \mathcal{F}\setminus\mathcal{F}_{\text{Seward}}$)}},
  coltitle=white, colbacktitle=rmOrange!70!black,
  fonttitle=\bfseries,
  left=8pt, right=8pt, top=5pt, bottom=5pt,
]
\textit{Inquirer:} \textit{``Doctor Seward, whom did your dear friend mimic in private before her illness worsened?''}

\smallskip
\textit{Options:}\\
\quad(A) a Saxon child\\
\quad(B) Mr.\ Swales\\
\quad(C) a Magyar woman\\
\quad(D) Mina Harker\\
\quad(E) ``I cannot answer this from my own knowledge.''

\smallskip
\textsc{Gold}~\textcolor{rmGreen!55!black}{\textbf{(E)}} \hfill
\textsc{\bookworld}~\textcolor{rmRed}{(B)}\,{\large\textcolor{rmRed}{$\times$}} \hfill
\textsc{\ours}~\textcolor{rmGreen!55!black}{(E)}\,{\large\textcolor{rmGreen!55!black}{$\checkmark$}}

\smallskip
\textit{Why:} Lucy records her decision to imitate Mina's diary writing in a private diary entry. At the narrative point represented by this item, the source passage provides no evidence that Seward has learned of this decision; the expected response is therefore refusal.
\end{tcolorbox}

\end{minipage}
\caption{Two example \kbfqa items from \emph{Dracula}, posed to Dr.\ John Seward.}
\label{fig:benchmark-examples}
\end{figure}

\begin{table}[!tp]
\centering
\small
\setlength{\tabcolsep}{4pt}
\renewcommand{\arraystretch}{1.05}
\resizebox{\textwidth}{!}{%
\begin{tabular}{llcccccccc}
\toprule
\textbf{Book} & \textbf{Language} & $n_{\krf}\!:\!n_{\kr}$ & \textbf{direct} & \textbf{\bookworld} & \textbf{naive RAG} & \textbf{RAPTOR} & \textbf{HippoRAG} & \textbf{ComoRAG} & \textbf{\ours} \\
\midrule
A Study in Scarlet           & English & 344:256 & 35.2 & 30.3 & 19.0 & 20.7 & 30.8 & 37.1 & \textbf{68.8} \\
Tom Sawyer                   & English & 142:118 & 20.5 & 22.3 & 11.5 & 8.5  & 13.2 & 32.0 & \textbf{63.5} \\
Treasure Island              & English & 161:90  & 28.3 & 32.5 & 10.8 & 10.7 & 37.3 & 28.7 & \textbf{65.1} \\
Around the World in 80 Days  & English & 485:342 & 38.1 & 36.5 & 11.0 & 12.2 & 11.6 & 36.3 & \textbf{73.7} \\
Dracula                      & English & 502:419 & 29.1 & 39.9 & 18.8 & 18.5 & 22.4 & 35.1 & \textbf{75.5} \\
Paradise Lost                & English & 62:66   & 29.1 & 13.0 & 5.6  & 0.0  & 2.9  & 27.9 & \textbf{73.0} \\
\midrule
Solaris                      & Chinese & 202:97  & 39.1 & 32.0 & 28.5 & 24.9 & 34.5 & 42.2 & \textbf{71.0} \\
Ball Lightning               & Chinese & 544:556 & 32.7 & 35.0 & 24.5 & 18.7 & 16.9 & 28.4 & \textbf{78.0} \\
\bottomrule
\end{tabular}}
\caption{Per-book \kbfqa composition and \kbf scores (\%). Bold indicates the highest value in each row.}
\label{tab:task1-by-book}
\end{table}

\begin{table}[!tp]
\centering
\small
\setlength{\tabcolsep}{6pt}
\renewcommand{\arraystretch}{1.05}
\begin{tabular}{lcccccccc}
\toprule
\textbf{Method} & \textbf{KA} & \textbf{BA} & \textbf{EE} & \textbf{PT} & \textbf{IM} & \textbf{AD} & \textbf{BC} & \textbf{AVG} \\
\midrule
\ours                 & \textbf{4.88} & \textbf{4.97} & \textbf{5.00} & \textbf{4.93} & \textbf{5.00} & 3.22          & \textbf{4.91} & \textbf{4.70} \\
w/o Episodic Layer    & 4.82          & 3.03          & 3.00          & 3.15          & 3.00          & 3.37          & 3.08          & 3.35 \\
w/o Personality Layer & 4.80          & 3.12          & 3.03          & 3.30          & 3.03          & 3.45          & 3.17          & 3.41 \\
w/o Semantic Layer    & 4.68          & 3.10          & 3.00          & 3.23          & 3.00          & \textbf{3.47} & 3.12          & 3.37 \\
\bottomrule
\end{tabular}
\caption{Supplementary CharacterBox scores for the narrative layer ablations on all 226 script-guided instances from eight novels. Bold indicates the highest score in each column.}
\label{tab:characterbox-ablation}
\end{table}

\FloatBarrier
\begin{table}[!tp]
\centering
\small
\setlength{\tabcolsep}{8pt}
\begin{tabular}{@{}p{0.97\textwidth}@{}}
\hline
\textbf{Self-Probe Prompt} \\
\hline
You are \{character\}. \\
\\
\textbf{\#\# Current situation} \\
\{current\_context\} \\
\\
\textbf{\#\# Recalled scenes from your own first-person perspective (Episodic Layer)} \\
\{scene\_context\} \\
\\
\textbf{\#\# Probes already issued in this query (when present)} \\
\{prior\_probes\} \\
\\
\textbf{\#\# Facts already retrieved (when present)} \\
\{retrieved\_facts\} \\
\\
\textbf{\#\# Task} \\
Step 1 --- Read the scenes above and use them to sharpen what the current situation is really asking. In one short sentence, in your own voice, state your refined understanding of what is being asked, anchored to specific names, events, or relations that the scenes literally surfaced. \\
\\
Step 2 --- Formulate one concrete \textbf{question} about the next factual detail to retrieve. The question must concern a specific name, object, event, relation, time, or place --- \emph{not} a vague theme. It must \textbf{not} semantically overlap with any probe already issued, and it must be a new angle that the scenes (or already-retrieved facts) make worth pulling on. \\
\\
\textbf{\#\# Output format} \\
Return a single JSON object, double quotes only, no markdown: \\
\{ \\
\hspace*{1.5em}``refined\_query'': ``\textit{one short sentence sharpening what is being asked}'', \\
\hspace*{1.5em}``self\_probe'': ``\textit{one concrete question for the next retrieval round}'' \\
\} \\
\hline
\end{tabular}
\caption{\textsc{Self-Probe} prompt (Algorithm~\ref{alg:reveriemem}, lines~2 and~8).}
\label{tab:prompt-self-probe}
\end{table}

\begin{table}[!tp]
\centering
\small
\setlength{\tabcolsep}{8pt}
\begin{tabular}{@{}p{0.97\textwidth}@{}}
\hline
\textbf{Pattern Selection Prompt} \\
\hline
You are \{character\}. \\
\\
\textbf{\#\# Current situation} \\
\{current\_context\} \\
\\
\textbf{\#\# Your current emotional state} \\
\{current\_emotion\} \\
\\
\textbf{\#\# Candidate behavioural patterns (Personality Layer)} \\
Below are \{num\_modes\} behavioural patterns distilled from your past behaviour --- recurring patterns in how you speak and act under different emotional arcs. Each pattern has an identifier and a short description: \\
\\
\{modes\_list\} \\
\\
\textbf{\#\# Task} \\
Which pattern best fits how you would naturally respond to the current situation, given your current emotional state? Pick \textbf{one}. If no pattern fits perfectly, pick the closest. \\
\\
\textbf{\#\# Output format} \\
Return a single JSON object, double quotes only, no markdown: \\
\{ \\
\hspace*{1.5em}``chosen\_mode\_id'': ``\textit{the identifier of the selected pattern}'', \\
\hspace*{1.5em}``brief\_reason'': ``\textit{one sentence explaining why this pattern fits}'' \\
\} \\
\hline
\end{tabular}
\caption{\textsc{Pattern} prompt (Algorithm~\ref{alg:reveriemem}, line~13).}
\label{tab:prompt-pattern}
\end{table}

\begin{table}[!tp]
\centering
\small
\setlength{\tabcolsep}{8pt}
\begin{tabular}{@{}p{0.97\textwidth}@{}}
\hline
\textbf{Memory-Fusion Prompt} \\
\hline
You are \{character\}. \\
\\
\textbf{\#\# Current situation} \\
\{current\_context\} \\
\\
\textbf{\#\# Scenes you have lived through (Episodic Layer, first-person)} \\
\{scene\_context\} \\
\\
\textbf{\#\# Specific facts your reasoning surfaced (Semantic Layer, visibility-bounded)} \\
\{retrieved\_facts\} \\
\\
\textbf{\#\# Reasoning conclusion grounded in these memories} \\
$\rho_c$ \\
\\
\textbf{\#\# Behavioural pattern conditioning your voice this turn (Personality Layer)} \\
- Description: \{pattern\_description\} \\
- Canonical excerpts: \{pattern\_samples\} \\
\\
\textbf{\#\# Task} \\
Sit with everything above end to end and respond to the current situation as \{character\} would. In 1--3 first-person sentences, meet the moment with what you actually recall --- name who, what, where, or when, with the certainty your recollection affords. If a small step is needed to go from what the facts literally show to what the situation asks, take that step visibly in your own voice. \\
\\
\textbf{\#\# Constraints} \\
1. Treat the scenes and the specific facts as \textbf{authoritative}; do not add detail they do not carry. \\
2. Restrict yourself to what \{character\} could plausibly know from these scenes and facts. If these scenes and facts do not support an answer, acknowledge the gap honestly in your own voice rather than fabricating a response. \\
3. Let the behavioural pattern shape your cadence, tone, and gesture: imitate the canonical excerpts' style, not their literal content. \\
4. Stay within \{character\}'s perspective; do not draw on knowledge beyond the supplied scene memories and visibility-allowed facts. \\
\\
\textbf{\#\# Output format} \\
Output the response directly --- first person, 1--3 sentences. No JSON, no preamble, no header. \\
\hline
\end{tabular}
\caption{\textsc{Memory-Fusion} prompt (Algorithm~\ref{alg:reveriemem}, line~17).}
\label{tab:prompt-memory-fusion}

\end{table}

\clearpage

\end{document}